\documentclass[letterpaper, 10 pt, conference]{ieeeconf}  % Comment this line out if you need a4paper

\DeclareUnicodeCharacter{202F}{\,}

\IEEEoverridecommandlockouts                              % This command is only needed if
\usepackage{graphicx} % for pdf, bitmapped graphics files
\usepackage[utf8]{inputenc}
\usepackage{amsmath,amsfonts,booktabs,cite} % general useful stuff
\usepackage{gensymb} % for \degree
\usepackage{siunitx,textcomp} % for the degree symbol
\usepackage{hyperref}
\usepackage{cleveref} % for easier referencing
\usepackage{subcaption}
\usepackage{tabularx}
\usepackage{makecell}
\usepackage{pbox}
\let\labelindent\relax
\usepackage[inline]{enumitem} % inline enumerate
\usepackage[nolist]{acronym} % acronyms by the \ac{label} command
\usepackage{multirow}
\usepackage[ruled,linesnumbered, vlined]{algorithm2e}
\usepackage{bm}
\usepackage{titlesec}
\usepackage{balance}
\usepackage[export]{adjustbox}
\usepackage{cellspace}
\usepackage{enumitem}
\usepackage{arydshln}  % For dashed lines in tables

\usepackage[font=small, justification=justified]{caption}

\def\figref#1{{Fig.~\ref{#1}}}
\def\tabref#1{{Tab.~\ref{#1}}}
\def\eqref#1{Eq.~(\ref{#1})}

\newcommand\etal{~\emph{et al. }}

\newsavebox{\twosubbox}

\usepackage{tikz}
\usetikzlibrary{arrows}
\usetikzlibrary{positioning,calc}
\usetikzlibrary{decorations.pathreplacing}
\usetikzlibrary{decorations.markings}
\usetikzlibrary{fit}
\usetikzlibrary{shapes.callouts}
\usetikzlibrary{shapes.geometric}
\usetikzlibrary{matrix}
\usetikzlibrary{arrows}
\usetikzlibrary{decorations.pathmorphing}
\usetikzlibrary{calligraphy}
\usetikzlibrary{shapes.arrows}
\usetikzlibrary{spy}
\usepackage{setspace}

\tikzset{
manip/.style={fill=Pastel2-A},
hardware/.style={fill=Pastel2-B},
perception/.style={fill=Pastel2-D},
system/.style={fill=Pastel2-C},
otg/.style={fill=Pastel2-B},
}

\pgfdeclarelayer{background}
\pgfdeclarelayer{foreground}
\pgfsetlayers{background,main,foreground}

\usepackage{scalefnt}

\usepackage{pgfplots}
\pgfplotsset{compat=1.14}
\usepgfplotslibrary{groupplots}
\usepgfplotslibrary{units}
\usepgfplotslibrary{statistics}
\usepgfplotslibrary{colorbrewer}

\usepackage{pgfplotstable}

\crefname{algocf}{alg.}{algs.}
\Crefname{algocf}{Algorithm}{Algorithms}

\titleclass{\subsubsubsection}{straight}[\subsection]

\newcounter{subsubsubsection}[subsubsection]
\renewcommand\thesubsubsubsection{\thesubsubsection.\arabic{subsubsubsection}}
\titleformat{\subsubsubsection}
{\normalfont\normalsize\bfseries}{\thesubsubsubsection}{1em}{}
\titlespacing*{\subsubsubsection}
{0pt}{3.25ex plus 1ex minus .2ex}{1.5ex plus .2ex}

\makeatletter
\renewcommand\paragraph{\@startsection{paragraph}{5}{\z@}%
	{3.25ex \@plus1ex \@minus.2ex}%
	{-1em}%
	{\normalfont\normalsize\bfseries}}
\renewcommand\subparagraph{\@startsection{subparagraph}{6}{\parindent}%
	{3.25ex \@plus1ex \@minus .2ex}%
	{-1em}%
	{\normalfont\normalsize\bfseries}}
\def\toclevel@subsubsubsection{4}
\def\toclevel@paragraph{5}
\def\toclevel@paragraph{6}
\def\l@subsubsubsection{\@dottedtocline{4}{7em}{4em}}
\def\l@paragraph{\@dottedtocline{5}{10em}{5em}}
\def\l@subparagraph{\@dottedtocline{6}{14em}{6em}}
\makeatother

\IfFileExists{graphicscache.sty}{\usepackage[dpi=400]{graphicscache}}

\newsavebox\CBox

\title{\LARGE \bf EvidMTL: Evidential Multi-Task Learning for Uncertainty-Aware Semantic Surface Mapping from Monocular RGB Images}
\author{Rohit Menon \and Nils Dengler \and Sicong Pan \and Gokul Krishna Chenchani \and Maren Bennewitz% <-this % stops a space
\thanks{All authors are with the Humanoid Robots Lab and the Center for Robotics, University of Bonn, Germany. Rohit Menon, Nils Dengler and Maren Bennewitz are additionally with the Lamarr Institute, Bonn, Germany.}
\thanks{This work has partially been funded by the Deutsche Forschungsgemeinschaft (DFG, German Research Foundation) under Germany’s Excellence Strategy, EXC-2070 -- 390732324 -- PhenoRob, by the DFG grant 459376902 – AID4Crops, and by the BMBF within the Robotics Institute Germany, grant No. 16ME0999.}
}%

\begin{document}
\maketitle
\thispagestyle{empty}
\pagestyle{empty}

%\def\thefootnote{*}\footnotetext{These authors contributed equally to this work.}
%change the footnote style back to arabic numbers
\renewcommand{\thefootnote}{\arabic{footnote}}

\begin{abstract}
For scene understanding in unstructured environments, an accurate and uncertainty-aware metric-semantic mapping is required to enable informed action selection by autonomous systems.
Existing mapping methods often suffer from overconfident semantic predictions, and sparse and noisy depth sensing, leading to inconsistent map representations.
In this paper, we therefore introduce EvidMTL, a multi-task learning framework that uses evidential heads for depth estimation and semantic segmentation, enabling uncertainty-aware inference from monocular RGB images.
To enable uncertainty-calibrated evidential multi-task learning, we propose a novel evidential depth loss function that jointly optimizes the belief strength of the depth prediction in conjunction with evidential segmentation loss.
Building on this, we present EvidKimera, an uncertainty-aware semantic surface mapping framework, which uses evidential depth and semantics prediction for improved 3D~metric-semantic consistency.
We train and evaluate EvidMTL on the NYUDepthV2 and assess its zero-shot performance on ScanNetV2, demonstrating superior uncertainty estimation compared to conventional approaches while maintaining comparable depth estimation and semantic segmentation.
In zero-shot mapping tests on ScanNetV2, \mbox{EvidKimera} outperforms Kimera by 30\% in semantic surface mapping accuracy and consistency, highlighting the benefits of uncertainty-aware mapping and underscoring its potential for real-world robotic applications.
\end{abstract}

\section{Introduction}\label{sec:intro}

A key aspect of robotic systems is semantic scene understanding, as is enables intelligent interaction~\cite{garg2020semantics} in applications such as autonomous driving, agriculture, and household robotics.
However, for robots to operate reliably in unstructured environments, they should not only recognize objects and surfaces but also quantify the uncertainty in their scene understanding, as wrong predictions can lead to inconsistent world models and therefore unreliable decision-making.

For scene understanding and mapping, traditional frameworks initially focused on purely geometric representations~\cite{hornung2013octomap, oleynikova2017voxblox}, which construct spatial occupancy maps but lack semantic context.
More recent semantic Truncated Signed Distance Field~(TSDF) mapping methods~\cite{rosinol2020kimera} have enabled dense volumetric representations by propagating 2D~semantic labels into 3D space.
However, these methods often suffer from overconfident predictions, unreliable depth sensing, and ambiguous \mbox{2D-to-3D label} fusion, leading to inconsistent map representations~\cite{guo2017on}.
This motivates the need for uncertainty-aware methods that quantify confidence in both depth and semantics, allowing robots to make more informed decisions.
Hence, semantic and depth predictions must not only be more accurate but their uncertainties should also correlate to the actual errors.

Bayesian uncertainty estimation techniques, such as Monte Carlo dropout~\cite{gal2016dropout} and ensemble learning~\cite{lakshminarayanan2017simple}, have been explored for semantic segmentation but remain computationally expensive due to multiple forward passes or the need for separate network instances~\cite{sharma2023bayesian}.
Similarly, while monocular RGB-based depth estimation methods, either standalone~\cite{ming2021deep} or integrated into multi-task frameworks such as SwinMTL\cite{taghavi2024swinmtl}, help mitigate sparse and noisy depth sensing, they still suffer from overconfidence and unreliability in challenging conditions\cite{gasperini2023robust}.

\begin{figure}[t]
    \centering
    \includegraphics[width=\linewidth, trim=100 0 100 0, clip]{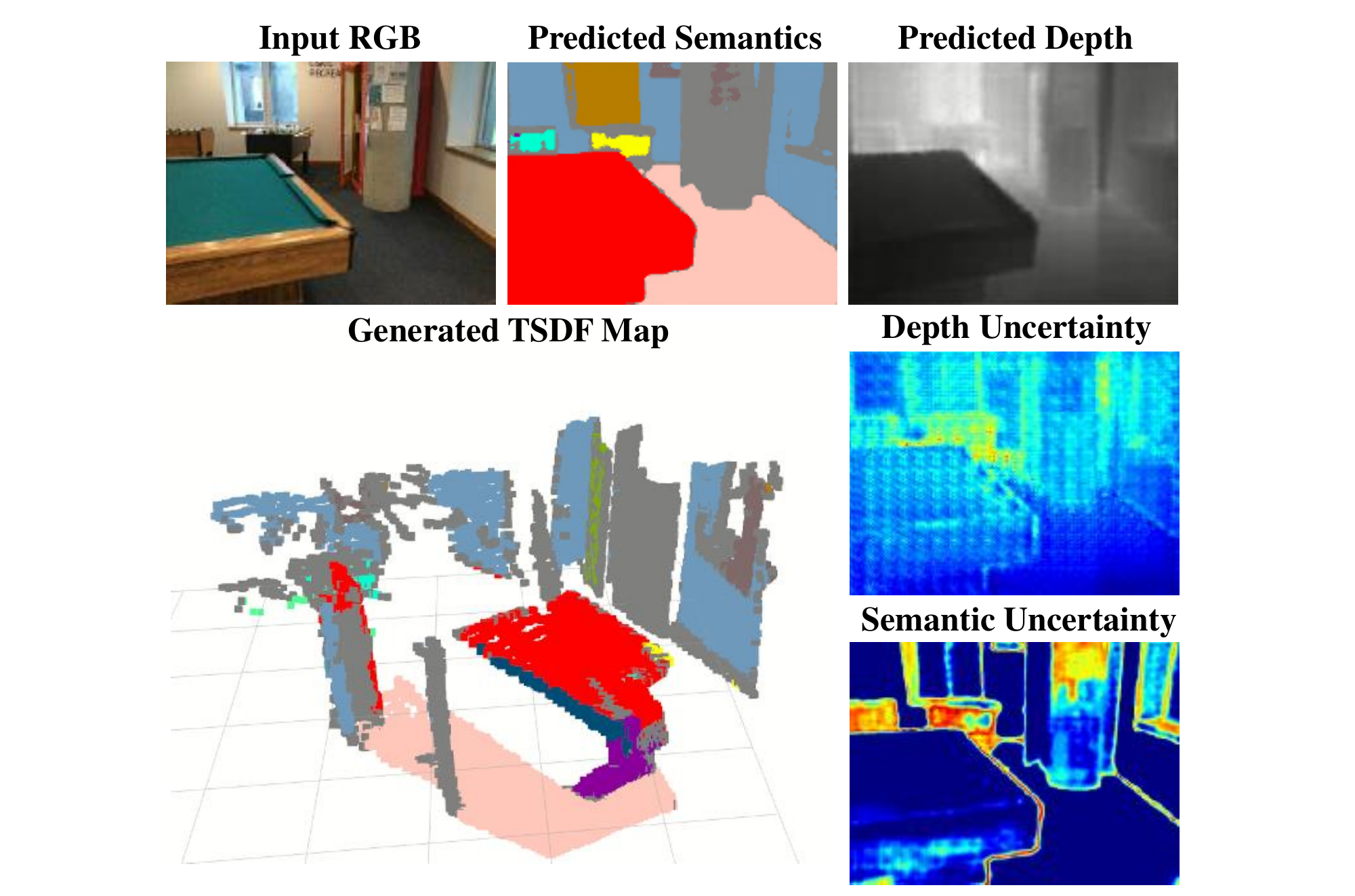}
    \caption{Visualization of our evidential multi-task perception pipeline. Given RGB image as input on top left, our EvidMTL framework predicts semantic labels and depth (top) along with their corresponding uncertainty estimates (right). The generated TSDF map, shown in bottom left, from our EvidKimera, leverages these uncertainty measurements, only including cells with low depth uncertainty and assigning unknown labels~(grey) to regions with high semantics uncertainty.}
    \label{fig:enter-label}
    \vspace{-5px}
\end{figure}

To address these limitations, we propose EvidMTL, an evidential multi-task learning framework that extends \mbox{SwinMTL~\cite{taghavi2024swinmtl}} with uncertainty-aware depth and semantic segmentation.
We propose a novel Evidential Scale-Invariant Log (EvidSiLog) loss, which integrates predictive uncertainty regularization with a novel prior-anchored Kullback–Leibler~(KL) divergence loss.
This KL divergence loss optimizes the hyperparameters of evidential depth prediction by anchoring them to the noise added ground-truth depth prior, with the predictive uncertainty regularized on the added noise, ensuring stable and effective joint learning of both tasks.
Next, we introduce \mbox{EvidKimera}, a semantic TSDF surface mapping framework that extends the multi-view fusion of Kimera~\cite{rosinol2020kimera} by integrating evidential predictions for depth and semantics.
EvidKimera employs a weighting strategy for \mbox{2D-to-3D} label transfer, discounting unreliable depth estimates to mitigate erroneous updates and incorporating viewpoint similarity to prevent the reinforcement of systematic errors.

To demonstrate the benefits of our loss design, we train and validate our networks on the NYUDepthV2~\cite{SilbermanECCV12} dataset and demonstrate its in-distribution performance.
Zero-shot testing on ScanNetV2~\cite{dai2017bundlefusion} shows that EvidMTL achieves superior uncertainty estimation compared to conventional approaches while maintaining comparable depth estimation and semantic segmentation performance.
To further explore the impact of uncertainty-aware semantic surface mapping, we conduct zero-shot mapping tests on ScanNetV2, confirming that EvidKimera outperforms Kimera in semantic surface mapping accuracy and consistency.
The code of our complete pipeline is available upon publication at \href{https://github.com/HumanoidsBonn/evidential_mapping}{\texttt{github.com/HumanoidsBonn/evidential\_mapping}}.

\section{Related Work}\label{sec:related_work}
\subsection{Joint Prediction of Semantic and Depth Information}
Dense semantic segmentation is well-established with convolutional architectures like U-Net~\cite{ronneberger2015u} and DeepLab~\cite{liang2015semantic}, while monocular depth estimation, pioneered by Eigen~\textit{et al.}~\cite{eigen2014depth} and later works such as Monodepth~\cite{monodepth17}, predicts dense depth maps from RGB images to address sparse sensing. Jointly learning these modalities reduces computational costs and improves robustness through inter-dependency.

Recent advances favor transformer-based architectures~(e.g., Swin Transformer~\cite{liu2022swin} for segmentation, and AdaBins~\cite{bhat2021adabins} for depth), enhancing global context but increasing complexity. Multi-task learning~(MTL) leverages shared representations to improve both tasks, as shown by SwinMTL~\cite{taghavi2024swinmtl}, which uses a shared transformer encoder with task-specific heads for efficiency.

Classical models lack uncertainty quantification, critical for safety-sensitive applications. Bayesian methods like Monte Carlo dropout~\cite{gal2016dropout} and deep ensembles~\cite{lakshminarayanan2017simple} provide uncertainty estimates but require multiple forward passes or mutliple models, limiting practicality.
To overcome these challenges, evidential approaches have been proposed to estimate uncertainty in a single pass for classification~\cite{sensoy2018evidential} and regression tasks~\cite{amini2020deep} respectively.

While Kendall\etal\cite{kendall2018multi} addressed multi-task loss balancing via homoscedastic uncertainty, recent work such as EMUFormer~\cite{landgraf2024efficient} employed a student-teacher distillation strategy to predict total uncertainty for joint segmentation and depth estimation. However, it lacks explicit decomposition into epistemic and aleatoric components and relies on a pretrained ensemble teacher. To our knowledge, no prior work incorporates evidential uncertainty into multi-task learning for this setting. We address this gap by introducing an evidential MTL framework that yields decomposable uncertainties while achieving state-of-the-art performance.

\subsection{Semantic Mapping}
By integrating 3D semantic information derived from 2D images, semantic mapping extends traditional metric maps~\mbox{\cite{hornung2013octomap, oleynikova2017voxblox}}.
Many existing approaches such as Kimera Semantics~\cite{rosinol2020kimera, grinvald2019volumetric}, employ Truncated Signed Distance Fields (TSDFs) for dense volumetric mapping.
However, they assign semantic labels using majority voting over hard labels from 2D projections, which limits the reliability of the final 3D map.

An alternative approach incorporates raw segmentation logits to represent class probabilities~\cite{sunderhauf2022meaningful}. While this method improves expressiveness, it still suffers from overconfident predictions and lacks a principled way to estimate uncertainty.
To mitigate this issue, Bayesian fusion has been explored~\cite{morilla2023robust}, but it relies on probabilistic neural networks, which introduce significant computational overhead.

The most relevant works to ours are those of Gan\etal\cite{gan2020bayesian}, Kim\etal\cite{kim2024evidential}, and Marques\etal\cite{marques2025mapspacebeliefprediction}, which all model semantic states using Dirichlet concentration parameters.
Gan\etal convert one-hot labels from classical segmentation into Dirichlet distributions, ignoring measurement uncertainty and thus lacking a true evidential foundation.
Kim\etal derive evidence from class probabilities of an evidential network, but discard measurement strength, leading to information loss.
Both use costly Bayesian Kernel Inference and infer occupancy from semantic posteriors.
Marques\etal predict 2D semantic belief states using hard labels and model 2D occupancy in evidential form with additional height maps, without supporting metric depth integration in 3D.
In contrast, our method performs occupancy-aware evidential integration that jointly updates semantic and TSDF weight and distance posteriors while preserving uncertainty, enabling robust 3D integration.

To the best of our knowledge, this is the first work to integrate evidential multi-task learning for uncertainty-aware semantic TSDF mapping from monocular images.
%Our approach leverages both depth uncertainty-based and viewpoint similarity-based discounting, enhancing fusion quality.

\section{Our Approach}\label{sec:methods}
\begin{figure*}[t]
\centering
\includegraphics[width=0.9\textwidth, trim= 0 0 0 0, clip]{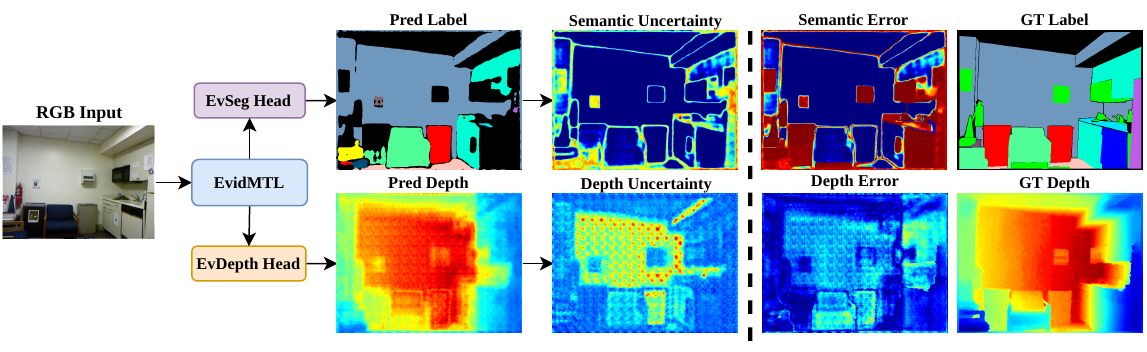}
\caption{From an input RGB image our proposed EvidMTL model jointly predicts semantics and depth estimates as well as their uncertainty~(left part of the figure).
The uncertainty estimates correspond well to the error in the prediction compared to the ground truth (GT) as shown on the right.}
\label{fig:evid_swinmtl}
\end{figure*}

%\subsection{Overview of Our Approach}

To enable reliable 3D scene reconstruction with uncertainty estimation, we propose evidential multi-task learning for depth and semantic predictions from monocular RGB images, along with an uncertainty-aware semantic surface mapping method.

\subsection{Evidential Multi-Task Learning}
For simultaneous semantics and depth prediction, we present \textbf{EvidMTL}, an evidential multi-task framework that concurrently predicts semantic labels and depth estimates while explicitly modeling uncertainty, all with a single pass using a shared encoder-decoder architecture, to generate inputs for the proposed mapping approach.

Building upon SwinMTL~\cite{taghavi2024swinmtl}, our framework leverages a Swin Transformer~\cite{liu2022swin} for multi-task learning.
This hierarchical vision transformer uses shifted windows to enable efficient self-attention, allowing for joint depth estimation and semantic segmentation.
In addition, our framework explicitly models evidential uncertainty for both tasks~(semantics and depth) and employs a tailored training scheme that mitigates gradient conflicts caused by the evidential regularization losses.
Thus, in contrast to SwinMTL, our approach enables stable multi-task learning and robust, uncertainty-aware predictions for subsequent semantic mapping.
Fig.~\ref{fig:evid_swinmtl} shows that from a single RGB input, our EvidMTL model jointly generates not only semantic segmentation and depth prediction but also their uncertainty estimates. The semantic and depth uncertainty correlate with the semantic and depth error respectively. This enables us to generate calibrated uncertainty-aware semantic maps.
%In contrast, our framework explicitly models evidential uncertainty for both tasks and employs a tailored training scheme that mitigates gradient conflicts caused by the evidential regularization losses.

\subsubsection{Evidential Depth Prediction} To predict the depth observation from RGB input, we assume a Gaussian distribution~\cite{belhedi2015noise} and model the predicted depth \(\mu\) with a Gaussian prior, placing its conjugate prior, the Normal Inverse Gamma (NIG) distribution, on the \mbox{variance \(\sigma^2\)}.
We replace SwinMTL's depth prediction head with an evidential regression head~\cite{amini2020deep} %Conv2NormalGamma layer~\cite{amini2020deep},
to generate the evidential depth parameters from the shared decoder.
Thus, for each pixel, instead of predicting only the expected depth~\(\mu\), our evidential depth regression head additionally outputs the hyper-parameters of the NIG distribution $\bigl[\alpha, \beta, \nu\bigr]$, where~\(\alpha\) quantifies confidence in the expected depth, \(\beta\) captures uncertainty in the depth noise, and \(\nu\) represents the evidence strength or virtual observation counts for \(\mu\).
The expected depth $\mathbb{E}[d]$, expected variance $\mathbb{E}[\sigma^2]$, and variance in expected depth $\mathit{Var}[d]$ are given as follows~\cite{amini2020deep}:
\begin{equation}
\mathbb{E}[d] = \mu, \quad
\mathbb{E}[\sigma^2] = \frac{\beta}{\alpha - 1}, \quad
\mathit{Var}[d] = \frac{\beta}{\nu(\alpha -1)}
\label{eq:expected_depth_variance}
\end{equation}
Here, \(\mathbb{E}[\sigma^2]\) represents the aleatoric uncertainty in the depth prediction~\(\mathit{u}^{\mathit{d}}_\mathit{al}\), which is irreducible and attributed to the data, while \(\mathit{Var}[d]\) represents the epistemic uncertainty~\(\mathit{u}^{\mathit{d}}_\mathit{ep}\), which reflects model uncertainty. The total modeled variance or predictive uncertainty is given by $\sigma²_{t} =\mathbb{E}[\sigma^2] + \mathit{Var}[d]$.

We extend the SwinMTL framework by modifying its depth loss to incorporate evidential hyper-parameters.
Our novel evidential depth loss \(\mathbb{L}_{ed}\) is defined as:
\begin{equation}
\label{eq:depth_loss}
\resizebox{0.91\columnwidth}{!}{$
\begin{aligned}
    &\mathbb{L}_\mathit{silog} &=& \hspace{0.5em} \sqrt{\mathbb{E} \left[ (\log\mathit{d}_\mathit{gt} - \log \mu)^2 \right] - \lambda \mathbb{E}[\log \mathit{d}_\mathit{gt} - \log \mu]^2}\\
    &\mathbb{L}_\mathit{unc} &=& \hspace{0.5em}\mathbb{E} \left[ \log(\sigma^2_{\text{t}}) - \log\left((\mathit{d}_\mathit{gt} - \mu)^2\right) \right] \\
    &\mathbb{L}_\mathit{reg} &=& \hspace{0.5em} \mathbb{D}_\mathit{KL}\left(\mathrm{NIG_\mathit{pred}}) \,\|\, (\mathrm{NIG_\mathit{prior}}) \right)\ \\
    &\mathbb{L}_\mathit{ed} &=& \hspace{0.5em} \mathbb{L}_\mathit{silog} +\min(1.0, (\frac{\mathit{n}^{\mathit{ep}}_\mathit{cur}}{\mathit{k} \cdot \mathit{n}^{\mathit{ep}}_\mathit{tot}})^2) (\lambda_1\cdot\mathbb{L}_\mathit{unc} +  \lambda_2\cdot \mathbb{L}_\mathit{reg})
\end{aligned}
$}
\end{equation}
\noindent Here, \(\mathbb{L}_\mathit{silog}\) is the Scale-Invariant Log (SiLog) loss~\cite{taghavi2024swinmtl}, \(\mathbb{L}_\mathit{unc}\) regularizes predictive uncertainty, and \(\mathbb{L}_\mathit{reg}\) is the Kullback–Leibler divergence loss~\cite{amini2020deep} between predicted hyper parameters $\mathrm{NIG_\mathit{pred}}$ and the prior NIG parameters $\mathrm{NIG_\mathit{prior}}$.
Additionally, \(\lambda_1\), \(\lambda_2\), and \(k\) are scaling coefficients, $\mathit{n}^{\mathit{ep}}_\mathit{cur}$ is the current epoch, and  $\mathit{n}^{\mathit{ep}}_\mathit{tot}$ is the total number of epochs.
To ensure stable learning, we introduce a square-law annealing for \(\mathbb{L}_\mathit{reg}\) and \(\mathbb{L}_\mathit{unc}\), gradually increasing its influence during training.
This prevents excessive regularization in early epochs while improving uncertainty-aware depth estimation.
Additionally, \(\mathbb{L}_\mathit{unc}\) enhances robustness by accounting for predictive uncertainty.

\subsubsection{Evidential Semantic Segmentation} In order to account for relative class probabilities and uncertainty in predictions, we model semantic segmentation, a multinomial classification task, as a Dirichlet distribution for evidential prediction.
Therefore, we extend SwinMTL’s~\cite{taghavi2024swinmtl} semantic segmentation head with an \textit{evidence layer} that transforms logits into class-specific evidence values using a softplus activation:
\begin{equation}
e_i = \operatorname{softplus}(z_i), \quad
c_i = e_i + 1
\end{equation}
where \( z_i \) is the logit for class \( i \), and \( c_i \) parametrizes the Dirichlet distribution.
The expected class probabilities and epistemic uncertainty are computed as:
\begin{equation}
\label{eq:sem_epi}
S = \sum c_i, \quad
p_i = \frac{c_i}{S}, \quad
u^{s}_\mathit{ep} = \frac{K}{S}
\end{equation}
where \(K\) is the number of semantic classes, \( S \) is the total evidence, and \( u^{s}_\mathit{ep} \) represents epistemic uncertainty, derived from Dempster-Shafer theory~\cite{shafer1992dempster}.
The total evidential semantic segmentation loss is defined as:
\begin{equation}
\resizebox{0.91\columnwidth}{!}{$
    \begin{aligned}
        &\mathbb{L}_\mathit{ece} &=&\hspace{0.5em} \sum^{\mathit{K}}_\mathit{k} \mathit{l}_k \cdot (\log S - \log \mathit{c}_\mathit{k}), \quad \mathbb{L}_\mathit{KL} \hspace{0.1em} = \hspace{0.1em}\mathbb{D}_\mathit{KL} \left( \mathit{Dir}(\mathbf{c}) \,\|\, \mathit{Dir}(\mathbf{1}) \right)\\
        &\mathbb{L}_\mathit{es}&=&\hspace{0.5em} \mathbb{L}_\mathit{ece} + \lambda_3\cdot\min\left(1.0, \frac{\mathit{n}^{\mathit{ep}}_\mathit{cur}}{\mathit{k} \cdot \mathit{n}^{\mathit{ep}}_\mathit{tot}}\right) \cdot \mathbb{L}_\mathit{KL}
    \end{aligned}
$}
\end{equation}
with \(\mathbb{L}_\mathit{ece}\) as the evidential cross-entropy loss~\cite{sensoy2018evidential}, \( \mathit{Dir}(\mathbf{1}) \) the uniform Dirichlet distribution, \( \mathit{Dir}(\mathbf{c}) \) the predicted distribution, and \( \mathbb{D}_\mathit{KL} \) the Kullback–Leibler divergence.
\(\mathbb{L}_\mathit{KL}\) acts as a regularizer, mitigating overconfident predictions.
This formulation ensures stable multi-task training by dynamically adjusting the strength of the regularization terms.
In particular, we apply evidential uncertainty modeling to SwinMTL’s semantic segmentation pipeline and employ linear annealing for semantic regularization to prevent conflicts with the squared-law depth regularization.

\subsection{Evidential Semantic Surface Mapping}

\begin{figure}[t]
    \centering
    \includegraphics[width=\linewidth, trim=70 100 65 125, clip]{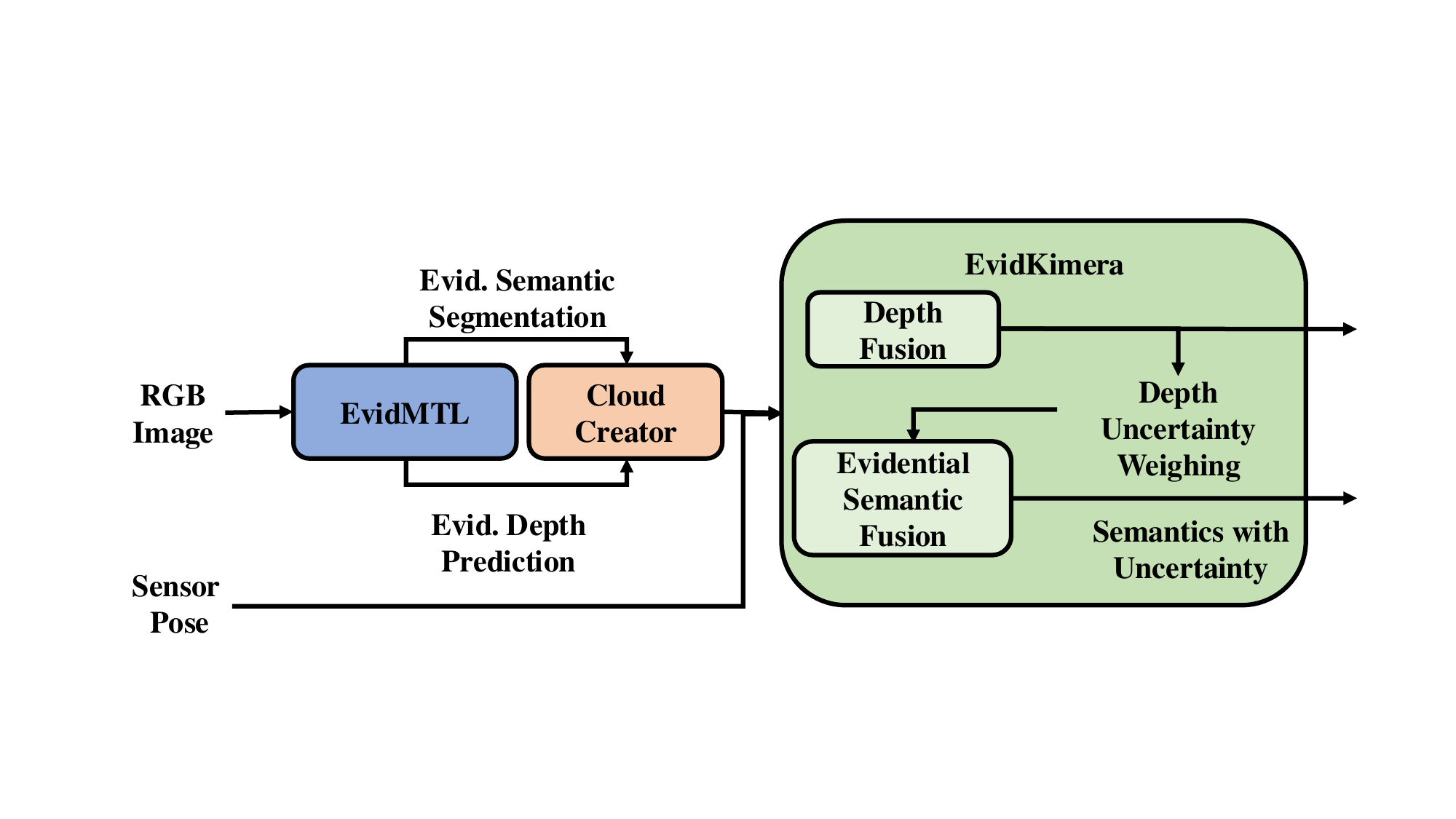}
    \caption{Our semantic evidential mapping framework: The RGB image is processed through EvidMTL (left), our evidential multi-task depth-semantic segmentation network.
The resulting semantic point cloud undergoes uncertainty-weighted Bayesian fusion for the TSDF layer, whereas the evidential semantic predictions are used as the measurements for updating the voxel semantic priors in EvidKimera (right).
The final uncertainty-weighted integration refines the semantic voxel posteriors.
The mapping framework outputs metric-semantic information with corresponding uncertainties.}
    \label{fig:sys_overview}
        % \vspace{-5px}

\end{figure}

\figref{fig:sys_overview} shows an overview of our proposed architecture and its three components.
It comprises three modules: (1) our \mbox{EvidMTL} network predicting depth and semantic segmentation with uncertainty, (2) a cloud creator fusing predictions into an evidential semantic point cloud, and (3) a mapping framework refining the global metric-semantic map via multi-view uncertainty-weighted integration.
The individual components are described in the following.

To combine the output of our evidential multi-task network into a meaningful map representation, we propose an uncertainty-aware semantic TSDF mapping framework that integrates an evidential semantic point cloud, formed by fusing the depth and semantic predictions:
\begin{equation}
\label{eq:ev_sem_cloud}
\resizebox{0.90\columnwidth}{!}{$
\mathcal{P} = \Bigl\{ p_i = \bigl( \mathbf{x}_i, \mathit{rgb}, {\mathit{u}{^{\mathit{d}}_\mathit{ep}}}_i, {\mathit{u}{^{\mathit{d}}_\mathit{al}}}_i, c_{i1}, \dots, c_{iK} \bigr) \;\Big|\; i=1,\dots,N \Bigr\}
$}
\end{equation}
where $\mathbf{x}_i = (x_i, y_i, z_i)$ denotes 3D coordinates, $\mathit{rgb}$ represents color, and $z_i$ is the expected depth $\mu$ in the camera frame.

\subsubsection{Evidential Depth Integration} In comparison to traditional TSDF mapping frameworks such as Voxblox~\cite{oleynikova2017voxblox} and KinectFusion~\cite{newcombe2011kinectfusion}, that assign TSDF weights based on an inverse square law of the depth distance, we propose to incorporate uncertainty-aware weighting.
Specifically, we compute the total uncertainty and update the TSDF weights as:
\begin{equation}
\label{eq:tsdf_weight}
    \mathit{u^{\mathit{d}}_\mathit{tot}} = \mathit{u^{\mathit{d}}_\mathit{ep}} + \mathit{u^{\mathit{d}}_\mathit{al}}, \quad
    \mathit{w_{\mathit{m}}} = \frac{1}{\mathit{u^{\mathit{d}}_\mathit{tot}}}, \quad
    \mathit{w_\mathit{post}} = \mathit{w_\mathit{prior}} + \mathit{w_{\mathit{m}}}
\end{equation}
where \(\mathit{u^{\mathit{d}}_\mathit{tot}}\) represents the total depth uncertainty, \(\mathit{w_{\mathit{m}}}\) is the measurement weight, and \(\mathit{w_\mathit{post}}\) is the updated weight after incorporating the prior information.

In addition to updating TSDF weights, we also maintain a separate voxel-wise epistemic uncertainty.
This uncertainty is updated in a Bayesian manner, using the harmonic mean of the prior and measurement epistemic uncertainties:
\begin{equation}
    \frac{1}{\mathit{u}^{\mathit{post}}_\mathit{ep}} = \frac{1}{\mathit{u}^{\mathit{prior}}_\mathit{ep}} + \frac{1}{\mathit{u}^{\mathit{m}}_\mathit{ep}}
\end{equation}
This formulation ensures that the epistemic uncertainty is refined progressively as more observations are incorporated.
Unlike conventional TSDF frameworks that solely rely on depth confidence heuristics, our approach explicitly accounts for both aleatoric and epistemic uncertainties, leading to a more robust and uncertainty-aware reconstruction.
\subsubsection{Evidential Semantic Integration}
Conventional semantic mapping uses majority voting~\cite{zhao2014semantic} or Bayesian fusion~\cite{morilla2023robust}, converting between probabilities and distributions. Instead, we represent each voxel’s semantic state as a Dirichlet distribution, exploiting its conjugate‐prior property for end‐to‐end fusion of prior, measurement, and posterior without extra conversions~\cite{frigyik2010introduction}. This preserves uncertainty and accumulates evidence consistently.

We extend the $K$ semantic classes with two additional states, free space ($\mathit{F}$) and unknown ($\mathit{U}$), by augmenting the Dirichlet concentration vector to length $L=K+2$. Differentiating the 2D background class from the voxel‐level unknown class allows us to distinguish confident unknowns from genuine uncertainty.
Given a new measurement $\mathit{Dir}(\mathbf{c}^{\mathit{m}})$, the posterior update follows:
\begin{equation}
\mathbf{c}^{\mathit{post}} = \mathbf{c}^{\mathit{prior}} + \lambda \mathbf{e}^{\mathit{m}},
\end{equation}
where $\lambda$ controls the influence of the measurement relative to prior evidence, and $\mathbf{e}^{\mathit{m}} = (\mathbf{c}^{\mathit{m}} - 1)$ represents the evidential belief of the measurement.
The class probabilities and hard label assignment are computed as:
\begin{equation}
p_k = \frac{c_k}{S}, \quad S = \sum_{k=1}^{L} c_k.
\end{equation}
\begin{equation}
\hat{k} =
\begin{cases}
\arg\max_k c_k, & \text{if } u^{\mathit{s}}_{\mathit{ep}} < \tau \\
\mathit{U}, & \text{otherwise}.
\end{cases}
\end{equation}
where $\tau$ is the uncertainty threshold factor.
This formulation ensures a recursive evidential update, maintaining probabilistic structure while accumulating confidence from multiple observations.

\subsubsection{Depth Measurement Weighting} Reliable depth estimates are critical for accurate \mbox{2D-to-3D} semantic fusion.
Instead of assuming uniform confidence, we introduce an \textit{occupancy belief weighting} mechanism that conditions semantic updates on depth uncertainty, ensuring geometric consistency.

\textbf{Occupancy belief modeling:} Unlike prior approaches that treat geometric and semantic uncertainties separately, we explicitly incorporate TSDF-based occupancy confidence into semantic updates.
The occupancy belief $o$ is defined as:
\begin{equation}
\resizebox{0.87\columnwidth}{!}{$
o = w^{\mathit{m}} \cdot
\begin{cases}
1 - \exp(-|d^{\mathit{m}}|), &\text{if}  \hspace{0.5em} d^{\mathit{m}} > 0 \hspace{0.5em} (\text{free}) \\
\exp(-|d^{\mathit{m}}|), &\text{if}  \hspace{0.5em} d^{\mathit{m}} < 0 \hspace{0.5em} (\text{occupied})
\end{cases}
$}
\end{equation}
where $w^{\mathit{m}}$ is the TSDF weight, and $d^{\mathit{m}}$ is the signed distance function (SDF) value.
The depth weighting factor $\lambda_d$ is then computed as:
\begin{equation}
\lambda_{d} = \max(o, \epsilon)
\end{equation}
where $\epsilon = 0.01$, ensuring that low-confidence measurements contribute minimally.

\textbf{Handling free space and unknown regions:} To maintain consistency, semantic updates incorporate depth-aware adjustments:
\begin{equation}
\mathbf{e}^{\mathit{m}} =
\begin{cases}
\lambda_{d} , & d > 0 \quad (\text{free space}) \\
\mathbf{e}^{\mathit{m}}, & d \leq 0 \quad (\text{occupied space}) \\
\lambda_{d} , & w < \epsilon \quad (\text{unknown}).
\end{cases}
\end{equation}
By explicitly modeling free-space evidence and preserving uncertainty in unknown regions, this approach ensures robustness to measurement noise while preventing erroneous semantic updates.

\subsubsection{Viewpoint Similarity Discounting} To prevent systematic error accumulation, we weight semantic observations by a viewpoint‐dissimilarity factor~\cite{menon2023nbv}, so that measurements from similar views are discounted.

By integrating these two components into the weighting factor, our framework ensures a robust and uncertainty-aware semantic fusion that maintains consistency across multiple observations while preventing erroneous updates.

\section{Experimental Evaluation}\label{sec:experiments}
The experiments are designed to demonstrate that: (1)~The proposed combination of SiLog, KL$_{\mu}$, and uncertainty regularization loss functions in our EvidMTL framework facilitate the generation of calibrated uncertainty estimates.
(2) Our EvidMTL network achieves comparable depth and semantic prediction while delivering superior uncertainty estimation compared to a state-of-the-art baseline, particularly on out-of-distribution data.
(3) Our evidential mapping framework, EvidKimera, in conjunction with the outputs of our EvidMTL network, produces an accurate and uncertainty-aware semantic map compared to conventional mapping.

\subsection{EvidMTL Evaluation}

\subsubsection{\textbf{Metrics and Baseline Methods}}

To evaluate our proposed \textit{EvidMTL} model, we report the following metrics:
\vspace{1mm}
\begin{itemize}[leftmargin=*, itemsep=1pt]
  \item \textbf{mIoU}: Mean Intersection-over-Union for semantic segmentation.
  \item \textbf{Pixel Acc}: Pixel accuracy for semantic segmentation.
  \item \textbf{Seg ECE}: Expected Calibration Error; measures alignment between predicted semantic uncertainty (\eqref{eq:sem_epi}) and actual segmentation errors.
  \item \textbf{RMSE}: Root Mean Squared Error for depth prediction.
  \item \textbf{Depth NLL}: Negative Log-Likelihood; penalizes inaccurate or overconfident depth predictions.
  \item \textbf{Normalized Depth ECE}: Expected Calibration Error; measures alignment between predicted depth uncertainty~(\eqref{eq:expected_depth_variance}) and actual depth errors normalized by the maximum depth of 10m.
  \item \boldmath{$\nu$}\unboldmath: Strength of predicted depth evidence.
\end{itemize}
\subsubsection{\textbf{Training and Evaluation Setup}}
We train and validate our networks on the NYUDepthV2 dataset~\cite{SilbermanECCV12}, which comprises 795 training images and 694~validation images. All models were then trained on single NVIDIA RTX A6000 GPU with identical hyper-parameters: batch size 4, 500 epochs, learning rate linearly warmed from $2\times10^{-5}$ to $2\times10^{-4}$, layer‐wise decay factor of 0.9, and weight decay of 0.05. Out‐of‐distribution performance was assessed on the ScanNetV2 dataset~\cite{dai2017bundlefusion}. To harmonize semantic annotations, NYU40 labels were remapped to the ScanNet22 ontology, an extension of ScanNet20 with two additional classes for “other furniture” and “other structures”. For a fair comparison, we re‐implemented and retrained the SwinMTL baseline~\cite{taghavi2024swinmtl} on these ScanNet22 labels, employing the Swin V2 Base SMIM backbone~\cite{xie2023revealing,kim2022global}.
\subsubsection{\textbf{Evaluation Results on NYUDepthV2 Dataset}}
\begin{table*}[t]
	\centering
	\resizebox{\textwidth}{!}{
\begin{tabular}{rl|ccc|ccc|c}
\hline
\textbf{Type} &\textbf{Model}& \textbf{mIOU} $\uparrow$ & \textbf{Pixel Acc} $\uparrow$ & \textbf{Seg. ECE} $\downarrow$ & \textbf{RMSE} $\downarrow$  & \textbf{Depth NLL} $\downarrow$ & \textbf{Normalized Depth ECE} $\downarrow$ & $\mathbb{\nu}$  \\
\hline
SE &EvidSeg  &0.49&0.75&0.09& -- & -- & -- & -- \\
    SE &EvidDepth (\textbf{EvidSiLog+KL$_{\mu_{n}}$})  & -- & --& -- & 0.48 & 0.70 & 0.03 &1.23 \\
    SE &EvidDepth (NLL+Reg~\cite{amini2020deep}) & -- & -- & --    & 1.27 & 1.41 & 0.21 &0.16 \\
    \hline
    MN &SwinMTL~\cite{taghavi2024swinmtl}       & 0.48         & \textbf{0.77}  & --   & \textbf{0.46}  & --    & --    & -- \\
    \hline
    ME &NLL+Reg~\cite{amini2020deep} & 0.51          & 0.76  & 0.10 & 3.00  &2.65 & 0.21  & $2.0\times10^{-3}$ \\
    ME &NLL+KL~\cite{amini2020deep}                                 & 0.52          & 0.75  & 0.09 & 1.23  & 1.37  & 0.05  & 3.024 \\

    \hline
    ME &SiLog+Reg                          &\textbf{0.53}          & 0.77  & 0.08 & 0.47  &3.96  &2.08  &$1.0\times10^{-6}$ \\
    ME &\textbf{EvidSiLog}+Reg &0.52 &0.76 &0.09 &0.48 &2.43 &0.19 &$1.0\times10^{-3}$\\
    ME &SiLog+KL                           &0.53 &0.76  &0.09 &0.48  &3.05  &0.81  &31.61 \\
    ME &SiLog+\textbf{KL$_{\mu_n}$}           &0.53          &0.76  &0.09 &0.48  &1.57  &0.09  &31.91 \\
    ME &\textbf{EvidSiLog+KL$_{\mu}$} (Ours)  &0.51          & 0.76  & 0.09 & 0.48  &0.86  &0.05  & 293 \\
    ME &\textbf{EvidSiLog+KL$_{\mu_n}$} (Our EvidMTL)    & 0.52          & 0.76  & 0.08 & 0.48  & \textbf{0.68}  &\textbf{0.03}  &1.22 \\
    \hline
\end{tabular}
}
    \caption{Semantic and depth performance of the baseline \textbf{SwinMTL} and our network variants on the NYUDepthV2 validation split.
\textbf{S}=single‑task, \textbf{M} = multi‑task; \textbf{E}=evidential, \textbf{N}=normal (non‑evidential).
We evaluate eight distinct combinations of depth reconstruction terms (NLL, SiLog, EvidSiLog) and evidence regularizers (Reg, KL, KL$_\mu$, KL$_{\mu_n}$).
The evidence strength $\nu$ indicates how confidently the model explains aleatoric versus epistemic uncertainty.
Key observations:
(1) \emph{SiLog} consistently lowers RMSE relative to NLL.
(2) Only when SiLog loss is paired with the uncertainty loss and KL with priors, the models learns to adjust $\nu$ around the prior, yielding the best Depth‑ECE.
(3) Our final model, \textbf{EvidMTL} (EvidSiLog+KL$_{\mu_n}$), matches SwinMTL’s segmentation and depth accuracy while providing well‑calibrated depth uncertainty for no extra network cost.}

	\label{tab:ablation}
\end{table*}

\begin{table*}[t]
\centering
\resizebox{0.99\textwidth}{!}{
    \begin{tabular}{l|ccc|ccc} % Added '|' after 'Seg. ECE' for vertical separation
    \hline
    \textbf{Model} & \textbf{mIOU} $\uparrow$ & \textbf{Pixel Acc} $\uparrow$ & \textbf{Seg. ECE} $\downarrow$ & \textbf{RMSE} $\downarrow$ & \textbf{Depth NLL} $\downarrow$ & \textbf{Normalized Depth ECE} $\downarrow$ \\
    \hline
    SwinMTL~\cite{taghavi2024swinmtl}                    & 0.35  & 0.57  & --    &\textbf{0.42} & --   & -- \\
    SiLog+Reg              & 0.35  & 0.59  & \textbf{0.04}  & 0.43  & 1.09  & 0.28 \\
    SiLog+KL               & 0.36  & 0.59 &0.04 &0.43  &0.96  &0.22 \\
    SiLog+\textbf{KL$_{\mu_{n}}$} & 0.35  & 0.59 &0.04 &0.44  &0.78  &0.09 \\
    \textbf{EvidSiLog+KL$_{\mu}$} (Ours)   & 0.35  & 0.58  & 0.04  & 0.43  &1.73  &0.05 \\
    \textbf{EvidSiLog+KL$_{\mu_{n}}$} (Our EvidMTL) &\textbf{0.36} &0.59  &0.04  &0.42  & \textbf{0.72}  & \textbf{0.04} \\
    \hline
    \end{tabular}
}
\caption{Zero‑shot semantic and depth performance on \textbf{ScanNetV2} (out‑of‑distribution w.r.t.\ NYUDepthV2 training).  Bold numbers mark the best result in each column. Evidential losses with the noisy‑prior KL$_{\mu_n}$~term (last row) deliver the lowest Depth‑NLL and Depth‑ECE while retaining SwinMTL‑level segmentation and depth accuracy.}

\label{tab:scannet_v2}
\end{table*}
We keep the segmentation head and its cross‑entropy loss unchanged and vary only the depth branch.  Two main reconstruction terms are considered: the negative log‑likelihood~(\textbf{NLL}) of Amini\etal\cite{amini2020deep} and the scale‑invariant logarithmic loss (\textbf{SiLog}, Eq.~\ref{eq:depth_loss}).  Each can be augmented by one of two evidence regularisers: (i) the original Amini penalty (\textbf{Reg}), and (ii) a KL divergence to a conjugate prior with three flavours.  \textbf{KL} uses \emph{predicted} mean and $\beta$ together with weak hyper‑parameters ($\alpha{=}1.01,\nu{=}0.001$); \textbf{KL$_\mu$} adopts the \emph{ground‑truth} depth as prior mean, a fixed $\beta{=}0.1$, and strong hyper‑parameters ($\alpha{=}2.0,\nu{=}1.0$); \textbf{KL$_{\mu_n}$} employs the ground‑truth depth perturbed by Gaussian noise and recalculates $\beta$ from the expected variance, again with weak hyper‑parameters. Adding the predictive‑uncertainty penalty $\mathcal{L}_{\text{unc}}$ to SiLog is denoted \textbf{EvidSiLog}. Combining \{NLL, SiLog, EvidSiLog\} with \{Reg, KL, KL$_\mu$, KL$_{\mu_n}$\} yields eight multi‑task configurations; three single‑task ablations freeze one of the heads.  All models are trained on NYUDepthV2 with only 795 images, making uncertainty learning especially challenging.

The evaluation results are shown in Table~\ref{tab:ablation}. NLL‑based losses struggle with the small data regime, producing the worst depth accuracy (RMSE up to 3.00m) and calibration (Depth‑ECE 0.21) even after regularisation.  Switching to SiLog reduces RMSE to $\approx0.48$m, yet naïve regularisers either drive the model into extreme under‑confidence (SiLog+Reg, $\nu\!\sim\!10^{-6}$, Depth‑ECE 2.08) or over‑confidence (SiLog+KL, $\nu\!=\!31.6$, Depth‑ECE 0.81).  Imposing a \emph{noisy} ground‑truth prior (SiLog+KL$_{\mu_n}$) moderates evidence strength and cuts NLL from 3.05 to 1.57, but reliability improves decisively only when SiLog is paired with $\mathcal{L}_{\text{unc}}$.  The resulting \textbf{EvidSiLog+KL$_{\mu}$} and \textbf{EvidSiLog+KL$_{\mu_n}$} delivers the best calibrated depth uncertainty predictions while being comparable in mIOU and RMSE among multi‑task models. Although both models have similar total predictive uncertainty calibration, EvidSiLog+KL$_{\mu}$ is an over-confident model with high evidence strength whereas EvidSiLog+KL$_{\mu_n}$ is less confident in its depth prediction.

The single‑task EvidDepth (EvidSiLog+KL$_{\mu_n}$) shows no measurable benefit over the multi‑task counterpart, its NLL (0.70) and Depth‑ECE (0.03) are effectively identical to the multi‑task values (0.68 / 0.03). By retaining the same depth performance while adding only a lightweight segmentation head, the multi‑task configuration provides calibrated predictions for \emph{both} tasks at no additional cost.

\subsubsection{\textbf{Out-of-Distribution Testing on ScanNetV2 Dataset} }
\begin{figure}[b]
\centering
\includegraphics[width=1.0\linewidth, trim=0 0 0 0, clip]{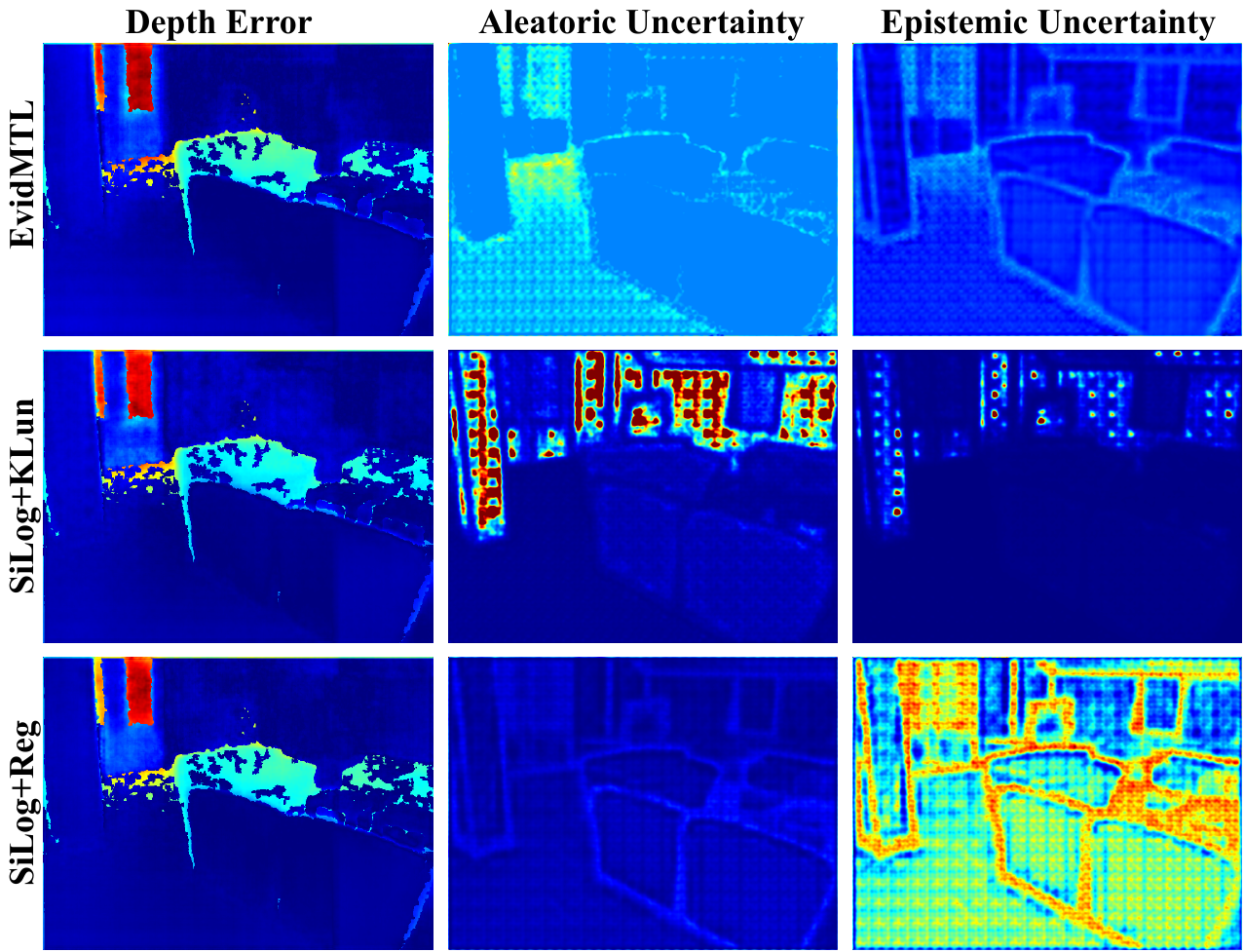}
\caption{
The columns show the depth error, aleatoric and epistemic uncertainty for a random scene from the ScanNetV2 dataset for the zero-shot evaluation of EvidMTL (EvidSiLog+KL$_{\mu_{n}}$), SiLog+KL$_{\mu_{n}}$, and SiLog+Reg. The error and uncertainty increase from blue to red. The high error red spots on the top are windows. SiLog+KL$_{\mu_{n}}$ (middle row) attributes errors to aleatoric uncertainty whereas SiLog+Reg (bottom row) attributes it to epistemic uncertainty. Our EvidMTL (top row) correctly shows epistemic uncertainty at object boundaries and aleatoric uncertainty on windows and low texture carpets on the floor.}
\label{fig:depth_error_uncertainty_correlation}
\end{figure}
To assess zero‑shot generalization, we evaluate \textbf{SwinMTL}, \textbf{SiLog+Reg}, \textbf{SiLog+KL}, \textbf{SiLog+KL$_\mu$+Unc}, \textbf{EvidSiLog+KL$_\mu$}, and our full \textbf{EvidSiLog+KL$_{\mu_n}$} on ten randomly selected scenes from the ScanNetV2 dataset~\cite{dai2017bundlefusion}. Since all models are trained exclusively on NYUDepthV2, ScanNetV2 introduces a significant domain shift while still containing indoor scenes.

The results in Table~\ref{tab:scannet_v2} show that:
(1) Our proposed EvidSiLog+KL$_{\mu_n}$ exceeds SwinMTL in segmentation accuracy (mIOU= 0.36, Pixel Acc = 0.59) and matches its lowest depth RMSE (0.42 m) while yielding the lowest Depth NLL (0.72) and Depth ECE (0.04).
(2) Incorporating the uncertainty loss with a noisy ground‑truth prior (KL$_{\mu_n}$) consistently calibrates depth uncertainty better than either the Reg or plain KL terms (Depth ECE drops from 0.28/0.22 to 0.04).
(3) Although SiLog+Reg and SiLog+KL retain competitive mIOU and RMSE, their higher NLL and ECE indicate unreliable confidence estimates, potentially problematic for downstream mapping.  Qualitative error‑versus‑uncertainty visualizations in Fig.~\ref{fig:depth_error_uncertainty_correlation} corroborate that EvidMTL produces the most plausible depth‑uncertainty maps under domain shift.
All models have an average inference time less than 100ms on an RTX3080Ti GPU.
In the next section, we explore the importance of introducing uncertainty for the network in the mapping task.
\subsection{Evidential Semantic Mapping Evaluation}
\begin{table*}[t]
  \centering
  \resizebox{0.99\textwidth}{!}{%
  \begin{tabular}{l|l|c|c|c|c|c|c}
    \hline
    \textbf{Framework} & \textbf{MTL Model} & \textbf{Voxel Repr.} & \textbf{Seg. Repr.} & \textbf{TSDF Weight} & \textbf{3D mIOU} $\uparrow$ & \textbf{Seg. Voxel Acc} $\uparrow$ & \textbf{Seg. Voxel ECE} $\downarrow$ \\
    \hline
    \multirow{2}{*}{Kimera}  & 2DGT (GT labels)                    & hard             & hard            & 1/d$^2$       & \textbf{0.54} & \textbf{0.69} & -- \\
                             & SwinMTL~\cite{taghavi2024swinmtl} & hard             & hard            & 1/d$^2$     & 0.22          & 0.36          & -- \\
    \hline
    %\multirow{1}{*}{ProbKimera}(Ours) & EvidMTL                       & prob             & evid$->$prob            & 1/$\mathit{u^{\mathit{d}}_\mathit{tot}}$      & --            & --            & -- \\
    %\hline
    \multirow{5}{*}{EvidKimera (Ours)} & SwinMTL ($u^s_{ep}<0.5$)              & evid             & logits$\to$evid            & 1/d$^2$     & 0.02            & 0.05            & 0.21 \\
    & SwinMTL   ($u^s_{ep}<0.5$)            & evid             & one-hot$\to$evid            & 1/d$^2$     & 0.01            & 0.02            & - \\
    & EvidMTL  ($u^s_{ep}<0.3$)               & evid             & evid            & 1/d$^2$     & 0.24            & 0.36            & 0.43 \\
                                        & EvidMTL ($u^s_{ep}<0.3$) & evid             & evid            & 1/$\mathit{u^{\mathit{d}}_\mathit{tot}}$          & 0.26          & 0.39          & \textbf{0.43} \\
                                        & EvidMTL ($u^s_{ep}<0.5$) & evid             & evid            & 1/$\mathit{u^{\mathit{d}}_\mathit{tot}}$          & 0.29          & 0.43          & 0.43 \\
    \hline
  \end{tabular}%
  }
  \caption{Zero-shot 3D semantic surface mapping evaluation on ScanNetV2. Columns denote mapping framework, multi-task learning (MTL) model, semantic representation used for voxels and pixels (hard, evidential), TSDF weighting (distance-based $1/d^2$ or uncertainty-based 1/$\mathit{u^{\mathit{d}}_\mathit{tot}}$), and the evaluation metrics.}
  \label{tab:mapping}
\end{table*}

\subsubsection{\textbf{Metrics and Baseline Methods}}
To evaluate our evidential semantic mapping framework, we use the following metrics to assess the quality of the generated semantic TSDF maps:
\vspace{1mm}
\begin{itemize}[leftmargin=*, itemsep=1pt]
  \item \textbf{3D mIoU}: Mean cumulative Intersection-over-Union across all scenes for voxels with ground truth correspondence within a threshold (equal to voxel size).
  \item \textbf{Segmentation Voxel Accuracy}: Fraction of voxels with correct semantic labels.
  \item \textbf{Segmentation Voxel ECE}: Expected Calibration Error; measures the correlation between predicted voxel uncertainty and semantic label error.
\end{itemize}
\vspace{1mm}
We compare the following mapping variants, as summarized in Table~\ref{tab:mapping}:
\vspace{1mm}
\begin{itemize}[leftmargin=*, itemsep=1pt]
  \item \textbf{Kimera + 2DGT}: Hard voxel and hard segmentation using ground‐truth labels; TSDF weight $1/d^{2}$.
  \item \textbf{Kimera + SwinMTL}: Hard voxel and hard segmentation via majority vote of SwinMTL predictions; TSDF weight $1/d^{2}$.
  \item \textbf{EvidKimera + SwinMTL-Logits}: Evidential fusion of SwinMTL logits (evid/logits$\to$ evid); TSDF weight $1/d^{2}$.
  \item \textbf{EvidKimera + SwinMTL-OneHot}: Evidential fusion of SwinMTL hard labels encoded as evidence (evid/one hot$\to$ evid); TSDF weight $1/d^{2}$.
  \item \textbf{EvidKimera + EvidMTL}: Full EvidMTL Dirichlet evidential fusion (evid/evid); TSDF weight $1/d^{2}$.
  \item \textbf{EvidKimera + EvidMTL ($u^s_{ep}<0.3$)}: Full EvidMTL Dirichlet evidential fusion with total‐depth‐uncertainty TSDF weighting ($1/\mathit{u^{\mathit{d}}_\mathit{tot}}$) and segmentation‐uncertainty threshold 0.3.
  \item \textbf{EvidKimera + EvidMTL ($u^s_{ep}<0.5$)}: Same as above with segmentation‐uncertainty threshold 0.5.
\end{itemize}
\vspace{1mm}
All experiments run online on Ubuntu 20.04 with ROS Noetic on an 11th-Gen Intel i9-11900K CPU and NVIDIA RTX 3080 Ti GPU, with EvidKimera running at 1 Hz. To mimic real application scenarios, we perform zero-shot inference without fine-tuning the networks.
% \subsubsection{\textbf{Metrics and Baseline Methods}}

\subsubsection{\textbf{Zero-shot Evaluation Results}}
Our zero‐shot mapping experiments (see \tabref{tab:mapping}) show that even an idealized pipeline using perfect 2D semantics and depth (“Kimera + 2DGT”) achieves only 0.54 3D mIoU and 69\% Segmentation Voxel Accuracy. We attribute this gap to inevitable 2D–to–3D projection errors arising from view‐to‐view inconsistencies and to the overconfident nature of hard labels, which cannot express uncertainty about occlusions or noisy depth. When replacing the ground‐truth semantics with the SwinMTL predictions without any uncertainty modelling (“Kimera + SwinMTL”), performance collapses (3D mIoU = 0.22, accuracy = 0.36), underscoring how overconfident but incorrect 2D predictions further degrade the 3D reconstruction.

Interpreting SwinMTL outputs as evidence, either by casting logits into a Dirichlet form or by converting hard one‐hot labels into pseudo‐evidence, yields no semantic recovery (3D mIoU = 0.02, accuracy = 0.04), despite reasonable calibration error (ECE = 0.21). This confirms that naive evidential encoding of 2D outputs does not compensate for model uncertainty. In contrast, our full EvidMTL Dirichlet fusion quantifies epistemic uncertainty at both pixel and voxel levels, recovering much of the lost performance (3D mIoU = 0.24, accuracy = 0.36) under distance‐based TSDF weighting. By further adopting uncertainty‐aware TSDF weights (1/\(u_{\mathrm{tot}}^{d}\)), we improve the results to 29\% mIoU and 43\% accuracy while maintaining a Seg. Voxel ECE of = 0.43 which is on par with state‐of‐the‐art 2D calibration techniques~\cite{marques2024overconfidence}. These findings demonstrate that principled evidential fusion, combined with uncertainty‐aware integration, substantially closes the gap to the ground‐truth upper bound and provides reliable confidence estimates for downstream tasks.

\section{Conclusion}
\label{sec:conclusion}
In this work, we introduce EvidMTL and EvidKimera, which integrate an evidential multi-task learning framework for uncertainty-aware semantic surface mapping from monocular images.
Our EvidMTL network, featuring two novel loss terms, jointly predicts semantic segmentation and depth estimation while explicitly modeling uncertainty, thereby enhancing prediction reliability and consistency.
Furthermore, our evidential semantic mapping framework EvidKimera leverages uncertainty quantification in both semantic and depth predictions to generate an uncertainty-aware semantic TSDF map.
Compared to baselines, EvidMTL achieves comparable performance in depth and semantic prediction while providing superior uncertainty estimation, particularly in depth uncertainty estimation, which in turn boosts 3D mapping performance EvidKimera framework.
To our knowledge, this is the first evidential multi-task learning framework for semantic TSDF mapping.
\bibliographystyle{IEEEtran}
\balance
\bibliography{refs}

\begin{thebibliography}{10}
\providecommand{\url}[1]{#1}
\csname url@rmstyle\endcsname
\providecommand{\newblock}{\relax}
\providecommand{\bibinfo}[2]{#2}
\providecommand\BIBentrySTDinterwordspacing{\spaceskip=0pt\relax}
\providecommand\BIBentryALTinterwordstretchfactor{4}
\providecommand\BIBentryALTinterwordspacing{\spaceskip=\fontdimen2\font plus
\BIBentryALTinterwordstretchfactor\fontdimen3\font minus
  \fontdimen4\font\relax}
\providecommand\BIBforeignlanguage[2]{{%
\expandafter\ifx\csname l@#1\endcsname\relax
\typeout{** WARNING: IEEEtran.bst: No hyphenation pattern has been}%
\typeout{** loaded for the language `#1'. Using the pattern for}%
\typeout{** the default language instead.}%
\else
\language=\csname l@#1\endcsname
\fi
#2}}

\bibitem{garg2020semantics}
S.~Garg, N.~S{\"u}nderhauf, F.~Dayoub, D.~Morrison, A.~Cosgun, G.~Carneiro,
  Q.~Wu, T.-J. Chin, I.~Reid, S.~Gould, \emph{et~al.}, ``Semantics for robotic
  mapping, perception and interaction: A survey,'' \emph{Foundations and
  Trends{\textregistered} in Robotics}, 2020.

\bibitem{hornung2013octomap}
A.~Hornung, K.~M. Wurm, M.~Bennewitz, C.~Stachniss, and W.~Burgard, ``Octomap:
  An efficient probabilistic 3d mapping framework based on octrees,''
  \emph{Autonomous Robots}, 2013.

\bibitem{oleynikova2017voxblox}
H.~Oleynikova, Z.~Taylor, M.~Fehr, R.~Siegwart, and J.~Nieto, ``Voxblox:
  Incremental 3d euclidean signed distance fields for on-board mav planning,''
  in \emph{Proc.~of the IEEE/RSJ Intl.~Conf.~on Intelligent Robots and Systems
  (IROS)}, 2017.

\bibitem{rosinol2020kimera}
A.~Rosinol, M.~Abate, Y.~Chang, and L.~Carlone, ``Kimera: {A}n open-source
  library for real-time metric-semantic localization and mapping,'' in
  \emph{Proc.~of the IEEE Intl.~Conf.~on Robotics \& Automation (ICRA)}, 2020.

\bibitem{guo2017on}
C.~Guo, G.~Pleiss, Y.~Sun, and K.~Q. Weinberger, ``On calibration of modern
  neural networks,'' in \emph{Proc.~of the Intl.~Conf. ~on Machine Learning},
  ser. Proceedings of Machine Learning Research.\hskip 1em plus 0.5em minus
  0.4em\relax PMLR, 2017.

\bibitem{gal2016dropout}
Y.~Gal and Z.~Ghahramani, ``Dropout as a bayesian approximation: Representing
  model uncertainty in deep learning,'' in \emph{Proc.~of the Intl.~Conf. ~on
  Machine Learning}.\hskip 1em plus 0.5em minus 0.4em\relax PMLR, 2016.

\bibitem{lakshminarayanan2017simple}
B.~Lakshminarayanan, A.~Pritzel, and C.~Blundell, ``Simple and scalable
  predictive uncertainty estimation using deep ensembles,'' \emph{Advances in
  Neural Information Processing Systems}, 2017.

\bibitem{sharma2023bayesian}
M.~Sharma, S.~Farquhar, E.~Nalisnick, and T.~Rainforth, ``Do bayesian neural
  networks need to be fully stochastic?'' in \emph{Proc.~of the Intl.~Conf.~on
  Artificial Intelligence and Statistics (AIS)}.\hskip 1em plus 0.5em minus
  0.4em\relax PMLR, 2023.

\bibitem{ming2021deep}
Y.~Ming, X.~Meng, C.~Fan, and H.~Yu, ``Deep learning for monocular depth
  estimation: A review,'' \emph{Neurocomputing}, 2021.

\bibitem{taghavi2024swinmtl}
P.~Taghavi, R.~Langari, and G.~Pandey, ``Swin{MTL}: A shared architecture for
  simultaneous depth estimation and semantic segmentation from monocular camera
  images,'' in \emph{Proc.~of the IEEE/RSJ Intl.~Conf.~on Intelligent Robots
  and Systems (IROS)}, 2024.

\bibitem{gasperini2023robust}
S.~Gasperini, N.~Morbitzer, H.~Jung, N.~Navab, and F.~Tombari, ``Robust
  monocular depth estimation under challenging conditions,'' in \emph{Proc.~of
  the IEEE/CVF Conf.~on Computer Vision and Pattern Recognition (CVPR)}, 2023.

\bibitem{SilbermanECCV12}
P.~K. Nathan~Silberman, Derek~Hoiem and R.~Fergus, ``Indoor segmentation and
  support inference from rgbd images,'' in \emph{Proc.~of the Europ.~Conf.~on
  Computer Vision (ECCV)}, 2012.

\bibitem{dai2017bundlefusion}
A.~Dai, M.~Nie{\ss}ner, M.~Zoll{\"o}fer, S.~Izadi, and C.~Theobalt,
  ``Bundlefusion: Real-time globally consistent 3d reconstruction using
  on-the-fly surface re-integration,'' \emph{ACM Transactions on Graphics
  (TOG)}, 2017.

\bibitem{ronneberger2015u}
O.~Ronneberger, P.~Fischer, and T.~Brox, ``U-net: Convolutional networks for
  biomedical image segmentation,'' in \emph{Medical image computing and
  computer-assisted intervention--MICCAI 2015: 18th international
  conference}.\hskip 1em plus 0.5em minus 0.4em\relax Springer, 2015.

\bibitem{liang2015semantic}
C.~Liang-Chieh, G.~Papandreou, I.~Kokkinos, K.~Murphy, and A.~Yuille,
  ``Semantic image segmentation with deep convolutional nets and fully
  connected crfs,'' in \emph{Proc.~of the Intl.~Conf. ~on Learning
  Representations}, 2015.

\bibitem{eigen2014depth}
D.~Eigen, C.~Puhrsch, and R.~Fergus, ``Depth map prediction from a single image
  using a multi-scale deep network,'' \emph{Advances in Neural Information
  Processing Systems}, 2014.

\bibitem{monodepth17}
C.~Godard, O.~{Mac Aodha}, and G.~J. Brostow, ``Unsupervised monocular depth
  estimation with left-right consistency,'' in \emph{Proc.~of the IEEE/CVF
  Conf.~on Computer Vision and Pattern Recognition (CVPR)}, 2017.

\bibitem{liu2022swin}
Z.~Liu, H.~Hu, Y.~Lin, Z.~Yao, Z.~Xie, Y.~Wei, J.~Ning, Y.~Cao, Z.~Zhang,
  L.~Dong, \emph{et~al.}, ``Swin transformer v2: Scaling up capacity and
  resolution,'' in \emph{Proc.~of the IEEE/CVF Conf.~on Computer Vision and
  Pattern Recognition (CVPR)}, 2022.

\bibitem{bhat2021adabins}
S.~F. Bhat, I.~Alhashim, and P.~Wonka, ``Adabins: Depth estimation using
  adaptive bins,'' in \emph{Proc.~of the IEEE/CVF Conf.~on Computer Vision and
  Pattern Recognition (CVPR)}, 2021.

\bibitem{sensoy2018evidential}
M.~Sensoy, L.~Kaplan, and M.~Kandemir, ``Evidential deep learning to quantify
  classification uncertainty,'' \emph{Advances in Neural Information Processing
  Systems}, 2018.

\bibitem{amini2020deep}
A.~Amini, W.~Schwarting, A.~Soleimany, and D.~Rus, ``Deep evidential
  regression,'' \emph{Advances in Neural Information Processing Systems}, 2020.

\bibitem{kendall2018multi}
A.~Kendall, Y.~Gal, and R.~Cipolla, ``Multi-task learning using uncertainty to
  weigh losses for scene geometry and semantics,'' in \emph{Proc.~of the
  IEEE/CVF Conf.~on Computer Vision and Pattern Recognition (CVPR)}, 2018.

\bibitem{landgraf2024efficient}
S.~Landgraf, M.~Hillemann, T.~Kapler, and M.~Ulrich, ``Efficient multi-task
  uncertainties for joint semantic segmentation and monocular depth
  estimation,'' in \emph{DAGM German Conference on Pattern Recognition}.\hskip
  1em plus 0.5em minus 0.4em\relax Springer, 2024.

\bibitem{grinvald2019volumetric}
M.~Grinvald, F.~Furrer, T.~Novkovic, J.~J. Chung, C.~Cadena, R.~Siegwart, and
  J.~Nieto, ``Volumetric instance-aware semantic mapping and 3d object
  discovery,'' \emph{IEEE Robotics and Automation Letters (RA-L)}, 2019.

\bibitem{sunderhauf2022meaningful}
N.~Sünderhauf, T.~T. Pham, Y.~Latif, M.~Milford, and I.~Reid, ``Meaningful
  maps with object-oriented semantic mapping,'' in \emph{Proc.~of the IEEE/RSJ
  Intl.~Conf.~on Intelligent Robots and Systems (IROS)}, 2017.

\bibitem{morilla2023robust}
D.~Morilla-Cabello, L.~Mur-Labadia, R.~Martinez-Cantin, and E.~Montijano,
  ``Robust fusion for bayesian semantic mapping,'' in \emph{Proc.~of the
  IEEE/RSJ Intl.~Conf.~on Intelligent Robots and Systems (IROS)}, 2023.

\bibitem{gan2020bayesian}
L.~Gan, R.~Zhang, J.~W. Grizzle, R.~M. Eustice, and M.~Ghaffari, ``Bayesian
  spatial kernel smoothing for scalable dense semantic mapping,'' \emph{IEEE
  Robotics and Automation Letters (RA-L)}, 2020.

\bibitem{kim2024evidential}
J.~Kim, J.~Seo, and J.~Min, ``Evidential semantic mapping in off-road
  environments with uncertainty-aware bayesian kernel inference,'' in
  \emph{Proc.~of the IEEE/RSJ Intl.~Conf.~on Intelligent Robots and Systems
  (IROS)}, 2024.

\bibitem{marques2025mapspacebeliefprediction}
J.~M.~C. Marques, N.~Dengler, T.~Zaenker, J.~Mucke, S.~Wang, M.~Bennewitz, and
  K.~Hauser, ``Map space belief prediction for manipulation-enhanced mapping,''
  2025.

\bibitem{belhedi2015noise}
A.~Belhedi, A.~Bartoli, S.~Bourgeois, V.~Gay-Bellile, K.~Hamrouni, and P.~Sayd,
  ``Noise modelling in time-of-flight sensors with application to depth noise
  removal and uncertainty estimation in three-dimensional measurement,''
  \emph{IET Computer Vision}, 2015.

\bibitem{shafer1992dempster}
G.~Shafer, ``Dempster-shafer theory,'' \emph{Encyclopedia of artificial
  intelligence}, 1992.

\bibitem{newcombe2011kinectfusion}
R.~A. Newcombe, S.~Izadi, O.~Hilliges, D.~Molyneaux, D.~Kim, A.~J. Davison,
  P.~Kohi, J.~Shotton, S.~Hodges, and A.~Fitzgibbon, ``Kinectfusion: Real-time
  dense surface mapping and tracking,'' in \emph{2011 10th IEEE international
  symposium on mixed and augmented reality}.\hskip 1em plus 0.5em minus
  0.4em\relax IEEE, 2011.

\bibitem{zhao2014semantic}
Z.~Zhao and X.~Chen, ``Semantic mapping for object category and structural
  class,'' in \emph{Proc.~of the IEEE/RSJ Intl.~Conf.~on Intelligent Robots and
  Systems (IROS)}, 2014.

\bibitem{frigyik2010introduction}
B.~A. Frigyik, A.~Kapila, and M.~R. Gupta, ``Introduction to the dirichlet
  distribution and related processes,'' \emph{Department of Electrical
  Engineering, University of Washignton, UWEETR-2010-0006}, 2010.

\bibitem{menon2023nbv}
R.~Menon, T.~Zaenker, N.~Dengler, and M.~Bennewitz, ``Nbv-sc: Next best view
  planning based on shape completion for fruit mapping and reconstruction,'' in
  \emph{Proc.~of the IEEE/RSJ Intl.~Conf.~on Intelligent Robots and Systems
  (IROS)}, 2023.

\bibitem{xie2023revealing}
Z.~Xie, Z.~Geng, J.~Hu, Z.~Zhang, H.~Hu, and Y.~Cao, ``Revealing the dark
  secrets of masked image modeling,'' in \emph{Proc.~of the IEEE/CVF Conf.~on
  Computer Vision and Pattern Recognition (CVPR)}, 2023.

\bibitem{kim2022global}
D.~Kim, W.~Ka, P.~Ahn, D.~Joo, S.~Chun, and J.~Kim, ``Global-local path
  networks for monocular depth estimation with vertical cutdepth,'' arXiv
  preprint arXiv:2201.07436.

\bibitem{marques2024overconfidence}
J.~M.~C. Marques, A.~J. Zhai, S.~Wang, and K.~Hauser, ``On the overconfidence
  problem in semantic 3d mapping,'' in \emph{Proc.~of the IEEE Intl.~Conf.~on
  Robotics \& Automation (ICRA)}, 2024.

\end{thebibliography}
\end{document}